\documentclass[runningheads]{llncs}
\usepackage{graphicx}
\usepackage{comment}
\usepackage{amsmath,amssymb} 
\usepackage{color}

\usepackage[width=122mm,left=12mm,paperwidth=146mm,height=193mm,top=12mm,paperheight=217mm]{geometry}

\usepackage[ruled,vlined]{algorithm2e}

\usepackage{authblk}

\newcommand{\attn}[1]{\textcolor{red}{(#1)}}

\begin{document}
	
	\pagestyle{headings}
	\mainmatter

	\title{UniformAugment: A Search-free Probabilistic Data Augmentation Approach} 
	

	
	\titlerunning{UniformAugment}
	
	%
	\author{Tom Ching LingChen\inst{1}\footnote[1]{Contributed equally.} \and
		Ava Khonsari\inst{1 \star} \and
		Amirreza Lashkari\inst{1 \star} \and
		Mina Rafi Nazari\inst{1 \star} \and
		Jaspreet Singh Sambee\inst{1 \star} \and
		Mario A. Nascimento\inst{2}}
	
	\authorrunning{ }
	%
	
	\institute{\email{\{tom.chen1, ava.khonsari, amir.lashkari1, mina.rafinazari, jaspreet.singh.sambee\}@huawei.com} \\
		Distributed Data Lab, Huawei Technologies, Canada \and 
		\email{mario.nascimento@ualberta.ca} \\
		Dept. of Computing Science, University of Alberta, Canada}
	
	\maketitle
	
	\begin{abstract}
		%
		%
		Augmenting training datasets has been shown to improve the learning effectiveness for several computer vision tasks. A good augmentation produces an augmented dataset that adds variability while retaining the statistical properties of the original dataset. Some techniques, such as AutoAugment and Fast AutoAugment, have introduced a search phase to find a set of suitable augmentation policies for a given model and dataset.  This comes at the cost of great computational overhead, adding up to several thousand GPU hours. More recently RandAugment was proposed to substantially speedup the search phase by approximating the search space by a couple of hyperparameters, but still incurring non-negligible cost for tuning those. In this paper we show that, under the assumption that the augmentation space is approximately distribution invariant, a uniform sampling over the continuous space of augmentation transformations is sufficient to train highly effective models. Based on that result we propose UniformAugment, an automated data augmentation approach that completely avoids a search phase. In addition to discussing the theoretical underpinning supporting our approach, we also use the standard datasets, as well as established models for image classification, to show that UniformAugment's effectiveness is comparable to the aforementioned methods, while still being highly efficient by virtue of not requiring any search.
		
		
		\keywords{Data augmentation, image processing, image classification}
	\end{abstract}

	\section{Introduction}
	
	Data augmentation is an effective technique for training machine learning models, especially for image classification. Data augmentation virtually increases the dataset size in order to help regularizing the model, and increases the diversity of the data to help the model learn invariance.
	While this ideally leads to improvement in accuracy and generalization, increasing the diversity of the data requires some care. While, during training, we want the augmented data point to be different from the original one, we still want it to be a correct representative of the original label. For example, in image classification, if an image is cropped at an area that does not contain the main object, the augmented image is no longer representative of the original class. Therefore, to balance the diversity vs. correctness trade off, conventional data augmentations, such as cropping, blurring, etc, require domain expertise and manual work in order to deliver an effective augmentation strategy.
	
	Recent research has focused on automatically finding optimal data augmentations while avoiding manual design. Automated augmentation methods first assume augmentation space generated by a set of standard augmentation operations and their parameters, and then use a search algorithm to find a set of augmentation operations and their parameters within this space that reduce the loss of the model. However, searching through the augmentation space imposes substantial computational overhead. Column two of Table~\ref{table:summary} shows the size of the augmentation space for the current automatic augmentation methods: AutoAugment (AA) \cite{cubuk2019autoaugment}, Fast AutoAugment (FAA) \cite{lim2019fast}, Population Based Augmentation (PBA) \cite{ho2019population}, and RandAugment (RA) \cite{cubuk2019randaugment}. Moreover, in terms of GPU hours, AA and FAA require 15000 and 450 GPU hours respectively on ImageNet \cite{deng2009imagenet} dataset. Due to the high cost of searching algorithms, these methods do not scale to train large models and/or large datasets.

	\setlength{\tabcolsep}{4pt}
	\begin{table} [htb]
		\begin{center}
			\caption{
				Accuracy results for AA, FAA, PBA, RA and UA on CIFAR-10, CIFAR-100 \cite{krizhevsky2009learning} and ImageNet classification tasks. The models for the datasets are Shake-Shake 26 2x96d (SS) \cite{shake-shake}, Wide-ResNet-28-2 (WRN), and ResNet50 (RN50) \cite{resnet50}, respectively. Search space is the number of possible augmentations in the augmentation space. (N/A means no published result is available).}
			
			\label{table:summary}
			\begin{tabular}{l|c|c|c|c}
				\hline
				& Search & CIFAR-10 & CIFAR-100 & ImageNet\\
				& space & on SS & on WRN & on RN50\\
				\hline \hline 
				Baseline & 0    & 97.1 & 81.2  & 76.3\\
				\hline
				AA  & $10^{32}$ & 98.0 & 82.9  & 77.6 \\
				FAA & $10^{32}$ & 98.0 & 82.7  & 77.6 \\
				PBA & $10^{61}$ & 98.0 & 83.3  & N/A    \\
				RA  & $10^{2}$  & 98.0 & 83.3  & 77.6 \\
				UA & 0     & 98.1 & 82.8  & 77.7 \\
				\hline
			\end{tabular}
		\end{center}
	\end{table}
	
	In this work, we propose UniformAugment (UA), a \emph{search-free} data augmentation method that is efficient and scalable. We hypothesize that if the augmentation space is approximately invariant, that is, the majority of the augmented data remains representative of the original target label and within the distribution of the dataset, then searching for optimal augmentations within this space is not necessary.  In that case, uniformly sampling the augmentation operations from that augmentation space may be just as effective, thus forgoing the expensive cost of the search algorithm. 
	Our extensive experiments have shown that by using the same augmentation operations as presented in AA, but by sampling their parameters from a continuous space instead of a discrete space as done in AA's original proposal, we could achieve comparable results. 
	The last line in Table~\ref{table:summary} shows a summary of our results, compared to  methods mentioned above, using different datasets and models. 
	
	In addition to extensive experiments, we also analyzed our method and other automated augmentation methods using the theoretical framework proposed by Chen et al. \cite{chen2019invariance}. Using that framework, we show why uniform sampling is actually expected to achieve good results given an approximately invariant augmentation space. Besides being efficient and effective, UA can be easily implemented and applied to different models and tasks. One can view the design of UA as an instance of Occam’s Razor principle, i.e., one where the simplest solution is preferred. 
	
	The rest of the paper is organized as follows. In Section~\ref{sec: related work}, we present related works on automated data augmentations.
	Section~\ref{sec:UA} presents our main contribution, the UniformAugment algorithm, after discussing the theoretical analysis supporting it. Next, in Section~\ref{sec:experiments}, we compare our results with a baseline conventional data augmentation methods as well as AA, PBA, FAA, and RA, using standard datasets and models. Finally, Section~\ref{sec:conclusions} concludes the paper and offers directions for future work.
	
	\section{Related Work}
	\label{sec: related work}
	
	
	There has been extensive research on data augmentation for computer vision tasks. As mentioned earlier, traditional data augmentation methods require manual design and are usually dataset/domain dependent. For example random flip, and color distortions are commonly used for natural image datasets such as CIFAR-10 \cite{krizhevsky2009learning} and ImageNet \cite{deng2009imagenet}, while elastic distortions and scalings are more common on MNIST \cite{sato2015apac,simard2003best} dataset.
	%
	%
	Even though there is no universal agreement on which type of augmentations are always beneficial (or prejudicial), there is wide acknowledgement that designing an effective augmentation strategy for a model  requires domain knowledge and manual work.
	
	Recently there has been increased emphasis on \emph{searching} the optimal set of augmentation policies/strategies which includes a combination of augmentation functions applied to each sample during training. Recent advancements in such search-based methods, inspired by Neural Architecture Search (NAS) \cite{zoph2016neural}, gave rise to algorithms like AutoAugment (AA) \cite{cubuk2019autoaugment} which uses Reinforcement Learning (RL) to search for an optimal augmentation policy for a given task. They alternately train a child model and a Recurrent Neural Network (RNN) controller to learn the best policies on a proxy task and then use those policies for training the final model. Unfortunately this searching process requires extensive computing power, to the order of many thousands of GPU hours. 
	Fast AutoAugment (FAA) \cite{lim2019fast} utilizes Bayesian Optimization to find a data augmentation policy trained for density matching to improve the generalization ability of the model but this again requires an expensive separate search phase on a proxy task, in order of hundreds of GPU hours.
	
	With the augmentation search space defined by AA being widely used as a \emph{de facto} standard one, emphasis has been put on optimizing the complexity of  finding augmentation policies. Population Based Augmentation (PBA) \cite{ho2019population} aims to find an augmentation schedule by having multiple workers follow an ``exploit-and-explore'' procedure.  This approach discards the worst performing model and focuses on the better models while simultaneously exploring the search space of augmentation functions. Even though PBA is more time efficient in comparison to AA, it still uses proxy tasks and at the same time needs very large amounts of memory for the workers working concurrently. 
	
	Contrasting to all the approaches above, UA \emph{does not require any search} nor any hyperparameter tuning. Therefore, it is much more efficient than all those approaches and the results shown in Section~\ref{sec:experiments}, demonstrate that it is a just as effective augmentation method.
	
	In order to further reduce the computational complexity involved in the aforementioned methods, RandAugment (RA) \cite{cubuk2019randaugment} reduces the search space used in AA to 2 hyperparameters: number of operators and a single global distortion which can control the magnitude of \emph{all} augmentation operations at once. A set of augmentation operations are selected uniformly 
	and a grid-search is performed to find the optimal value of those hyperparameters.  
	A noteworthy conclusion in  \cite{cubuk2019randaugment} is that learning proxy tasks can provide sub-optimal results for a given task at hand. RA reduces the search space significantly but still needs extra computation to do grid search in order to find suitable hyperparameters. RA samples uniformly from a discrete search space, as introduced in AA, but we demonstrate in later sections that sampling uniformly in a continuous approximate invariant space can eliminate the need to search for any hyperparameters.

	
	There has also been several theoretical works on data augmentation. Rajput \emph{et al.} \cite{rajput2019does} analyzed how addictive spherical data augmentations affects the margin of classifiers. LeJeune \emph{et al.} \cite{lejeune2019implicit} proposed a Hessian-based rugosity measure as regularization penalty for training, and they showed the connection between data augmentation and regularization with this rugosity measure. Chapelle \emph{et al.} \cite{chapelle2001vicinal} proposed vicinity risk minimization principle that samples virtual training point out of the vicinity of Gaussian distribution centered at the actual training point. Zhang \emph{et al.} \cite{zhang2017mixup} uses this vicinity risk minimization principle to show how Mixup can help improve generalization. Finally, Chen \emph{et al.} \cite{chen2019invariance} use group invariance to show that data augmentation reduces the variance of loss and the gradient vector, which in turn reduces the empirical risk. In fact, the theoretical analysis supporting our approach (presented next) is largely based in that result.

	\section{UniformAugment}
	\label{sec:UA}
	
	
	In context of supervised learning, let $X$ be the data space, $Y$ the label space, and $\mathbb{P}$ the joint distribution of data and label where $(x,y) \sim \mathbb{P}$. We aim to find a mapping function $f^*$ between $X$ and $Y$ by estimating the parameters of function $f_{\theta}(X) \mapsto Y$ which minimizes a pre-defined loss function $l(f_{\theta}(x),y)$ over the data distribution $\mathbb{P}$. This is also known as minimizing the expected risk of function $f_{\theta}$, and it can be expressed as:
	\begin{align}
	f^* &= arg \min_{\theta} \int l(f_{\theta}(x),y)d \mathbb{P} (x,y)
	\end{align}
	
	Unfortunately, in most cases it is impossible to gain access to the true data distribution $\mathbb{P}$. In practice, we estimate $f^*$ by $\hat{f}$ which aims to minimize the Empirical Risk (ER) over a given data set $\{(x_i, y_i)\}_{i=1}^n$ which is assumed to mimic the data distribution $\mathbb{P}$. Hence, the minimization problem becomes:
	\begin{align}
	f^* \approx \hat{f} &= arg \min_{\theta} \frac{1}{n} \sum_{i=1}^{n} l(f_{\theta}(x_i),y_i)
	\end{align}
	While training, an augmentation transform $t \in T$ is applied to data point $(x_i, y_i)$, where $T$ is a set of approximate invariant augmentation transforms. From the Group Theory perspective, $T$ is a group of approximate invariant augmentation transforms acting on data set $\{(x_i, y_i)\}_{i=1}^n$ in a way that the augmented and original data have the approximate equality in distribution. With this in mind, an augmented data point $t(x_i)$, for all $t \in T$, form an orbit of element $x_i$. 
	
	Assuming augmentation transform $t \in T$ is sampled from the probability distribution $\mathbb{T}$, then the ER minimization equation can be re-written as \cite{chen2019invariance}:
	\begin{align}
	\hat{f} &= arg \min_{\theta} \frac{1}{n} \sum_{i=1}^{n} \int l(f_{\theta}(t(x_i)),y_i) d \mathbb{T} (t)
	\end{align}
	The integral in the equation above can be interpreted as the average loss of augmented data over group $T$. The averaging of the loss leads to reduction in variance of loss and gradient vector, which in turn tightens the ER bound between the empirical risk and the expected risk \cite{chen2019invariance}. If the gradient of loss varies significantly along the augmentation orbits of $x_i$, then the training procedure effectively denoises the gradient, which leads to a better training procedure. 
	
	Consider $T^{n} = T \times T \times ... \times T$ are the augmentations acting on the data set $\{(x_i, y_i)\}_{i=1}^n$ elementwise, as per \cite{chen2019invariance} the order of averaging and minimization can be reversed and the ER minimization estimator becomes:
	\begin{align}
	\hat{f} &= \mathbb{E}_{t_1,...,t_n \sim \mathbb{T}} \space arg \min_{\theta} \frac{1}{n} \sum_{i=1}^{n} l(f_{\theta}(t_i(x_i)),y_i)
	\end{align}
	
	In practice, this objective is minimized iteratively over a number of epochs with Stochastic Gradient Descent, so it is mandatory to find and apply the augmentation transforms with high variations epoch-wise. But, it is computationally expensive to re-do the search and calculate the above expectation every epoch, since it requires to calculate the gradient of loss for a high number of augmentation transform orbits. In order to avoid such a search, the augmentation transforms are uniformly sampled from the continuous approximate invariant augmentation space for all data points throughout the training procedure to cover the augmentation space uniformly. That is the main thrust supporting our proposed approach, UniformAugment (UA). In other words, UA assumes that the augmentation distribution $\mathbb{T}$ is a $\mathbb{U}$niform distribution over its parameters. With this alteration, the final ER minimization estimator is given by:
	\begin{align}
	\hat{f} &= \mathbb{E}_{t_1,...,t_n \sim \mathbb{U}} \space arg \min_{\theta} \frac{1}{n} \sum_{i=1}^{n} l(f_{\theta}(t_i(x_i)),y_i)
	\end{align}
	
	AA searches for augmentation policies that minimize the ER estimator on a proxy model with a subset of full data set and assumes that the found policies are transferable for training new model architectures with full data set. It has been shown that searching the proxy model to obtain the augmentation policies for a larger task is questionable \cite{cubuk2019randaugment}; and although transferring these policies increased the model accuracy, AA did not include a  comparison with a search-free approach to justify the search performance. FAA applies Bayesian Optimization search to find policies that minimize the loss on subsets of the data set in different folds. In other words, these augmentation policies aim to transform the unseen data points in a way similar to data set model is trained on which is not helpful for model generalization. Moreover, these policies minimize the loss of a trained model, so there is no guarantee that they minimize the loss over the training procedure. Unlike other approaches which search for an optimal set of policies for a specific data set, UA gives the model a chance to learn from the whole approximate invariant augmentation space by (1) uniformly sampling the augmentation transforms and (2) iteratively updating the model parameters with gradient descent. Finally, we argue that UA is also a suitable comparison to measure the  effectiveness of any augmentation search algorithm. 
	
	Given the above, UA's implementation is very simple as illustrated in Algorithm~\ref{alg:UA}. For each data point $(x,y)$ in the data set, multiple augmentation operations can be applied. Hyperparameter NumOps is the total number of augmentation transforms applied to the input data point in a sequential manner. Each augmentation transform is selected from augmentation transform set $T$. Since it is assumed that the distribution of augmentation transform set $T$ is Uniform, we uniformly sample an operation from the augmentation transform set. Each augmentation transform $t$ has two parameters, one is the probability of applying the transform $p$, and the other is the augmentation magnitude $\lambda \in [0,1] $ which determines the level of distortion. In UA, the distributions of $p$ and $\lambda$ are Uniform$(0,1)$ as well. At the end, each selected transform $t$ is applied to the input data $x$ with probability $p$ and magnitude $\lambda$. 
	A continuous augmentation parameter space covers a wider range of values and adds more variety to the data when compared to a discrete parameter space. The alternative, discretizing the augmentation parameter space, reduces the augmentation variance by limiting the space into a limited number of options for parameters.
	
	\begin{algorithm}[htb]
		\SetKwInOut{Input}{Input}
		\SetKwInOut{Return}{Return}
		\SetKwInOut{Set}{Set}
		\SetKwFunction{ApplyAugmentation}{ApplyAugmentation}
		\SetAlgoLined
		\Input{$(x,y) \in \{(x_i, y_i)\}_{i=1}^n$}
		\BlankLine
		\Set{$(\hat{x},\hat{y}) \gets (x,y)$}
		\For{$j \gets 1$ \KwTo $NumOps$}{
			$t_{ij} \gets t \in T \sim \mathbb{U}_{\{T\}} $\;
			$p_{ij} \gets p \sim \mathbb{U}_{(0,1)} $\;
			$\lambda_{ij} \gets \lambda \sim \mathbb{U}_{(0,1)} $\;
			$(\hat{x},\hat{y}) \gets t_{ij} (\hat{x},\hat{y}, p_{ij},\lambda_{ij} )$ \;
		}
		\Return{($\hat{x},\hat{y}$)}
		\caption{UniformAugment algorithm}
		\label{alg:UA}
	\end{algorithm}

	\section{Experiments}
	\label{sec:experiments}

	In the following we present the results of our experiments confirming that under the assumption that if the augmentation space is approximately invariant UA is capable of yielding training performance comparable to the approaches reviewed in Section~\ref{sec: related work}.
	
	We evaluate UA's performance on an image classification task using CIFAR10, CIFAR100 \cite{krizhevsky2009learning} and ImageNet \cite{deng2009imagenet} datasets. We compared UA results with those published by other recent methods, namely AA \cite{cubuk2019autoaugment}, FAA \cite{lim2019fast}, PBA \cite{ho2019population} and RA \cite{cubuk2019randaugment}\footnote{The missing experiments in those works are denoted by ``N/A'' in all tables that follow.}, 
	as well as Baselines which only include the default augmentations for of the following models: WideResNet~\cite{wideresnet}, Shake-Shake~\cite{shake-shake}, ResNet-50~\cite{resnet50} and ResNet-200~\cite{resnet200}.
	For a fair comparison we used the 15 augmentation functions  
	listed in Table~\ref{table:augrange} which were used by AA and other recent works.  The ranges for those operations are listed in the ``Default'' column of Table~\ref{table:augrange} as well. 
	Following the related work, each augmentation transform consists of two consecutive augmentation operations, i.e., the hyperparameter NumOps in Algorithm~\ref{alg:UA} is set to 2.
	%
	Finally, the training hyperparameters used for the classification models are the same as in \cite{lim2019fast} and we report the results averaged over 5 runs.
	
	Recall that, contrary to all of the competitors, UA requires no search phase, therefore in terms of efficiency, i.e., computing time, a comparison is not meaningful as UA is vastly superior to those methods.
	
	\subsubsection{CIFAR10.}
	Table~\ref{table:cifar10} shows the results of our experiments using UA on Wide-ResNet-40-2 (WRN-40-2), Wide-ResNet-28-10 (WRN-28-10) \cite{wideresnet} and Shake-Shake (SS) \cite{shake-shake} models. UA's accuracy is between 1-1.8\% better than the Baseline and $\sim0.5\%$ higher than CutOut augmentation. More importantly, the results are very much comparable to those by AA, PBA and FAA, namely around 0.1\%, with zero time spent on search phase, which is a constant w.r.t. UA in all experiments that follow.
	
	\begin{table} [htb]
		\begin{center}
			\caption{Average error rate of different augmentation methods on CIFAR10}
			\label{table:cifar10}
			\begin{tabular}{l | cccccc|l}
				\hline
				Model & Baseline & Cutout & AA & PBA  & FAA  & RA  & {UA}\\
				\hline \hline 
				WRN-40-2  & 5.6  & 4.1 & 3.7 & N/A & 3.6 & N/A &{3.75}\\
				WRN-28-10 & 3.9 & 3.1 & 2.6 & 2.6 & 2.7 & 2.7 & {2.67}\\
				SS (26 2x32d) & 3.6 & 3 & 2.5 & 2.5 & 2.5 & N/A &{2.49} \\
				SS (26 2x96d) & 2.9 & 2.6 & 2 & 2 & 2 & 2 & {1.90}\\
				\hline
			\end{tabular}
		\end{center}
	\end{table}
	
	\subsubsection{CIFAR100.}
	We also tested UA on CIFAR100 for Wide-ResNet-40-2 (WRN-40-2), Wide-ResNet-28-10 (WRN-28-10) 
	and Shake-Shake (SS). 
	Table \ref{table:headings2} shows 5-6\%, 
	improvement for Wide-ResNet-40-2, and 1-2\% improvement for Wide-ResNet-28-10  and Shake-Shake models comparing to the Baseline and CutOut. The error rates for UA are, as it was the case for CIFAR10, very much on par with AA, PBA, RA and FAA. 
	
	\begin{table} [htb]
		\begin{center}
			\caption{Average error rate of different augmentation methods on CIFAR100}
			\label{table:headings2}
			\begin{tabular}{l | cccccc|cl}
				\hline
				Model & Baseline & Cutout  & AA  & PBA  & FAA  & RA  & {UA}\\
				\hline \hline 
				\small WRN-40-2  & 26  & 25.2 & 20.7 & N/A & 20.6 & N/A & {20.99}\\
				\small WRN28-10 & 18.8 & 18.4 & 17.1 & 16.7 & 17.3 & 16.7 & {17.18}\\
				\small SS(26 2x96d) & 17.1 & 16 & 14.3 & 15.3 & 14.6 & N/A & {15.01}\\
				\hline
			\end{tabular}
		\end{center}
	\end{table}

	\subsubsection{ImageNet.}
	Finally, we tested UA on ImageNet for ResNet-50 \cite{resnet50} and ResNet-200 \cite{resnet200} models. Table \ref{table:headings3} shows classification error rates of UA comparing to Baseline, AA, FAA and RA methods. The error rates are 1.3\% and 1.9\% lower than the Baseline for ResNet-50 and ResNet-200 respectively.  As expected, once more, UA results are not only comparable, but are slightly better than AA's. 
	
	\begin{table} [htb]
		\begin{center}
			\caption{Average error rate of different augmentation methods on ImageNet}
			\label{table:headings3}
			\begin{tabular}{l | cccc|cl}
				\hline
				Model & Baseline & AA  & FAA  & RA & {UA}\\
				\hline \hline 
				ResNet50  & 23.7  & 22.4 & 22.4  & 22.4 & {22.37}\\
				ResNet200  & 21.5  & 20.0 & 19.4 & N/A & {19.6}\\
				\hline
			\end{tabular}
		\end{center}
	\end{table}

	\subsection{Investigating the ``approximately invariant augmentation space'' hypothesis}
	
	UA is based on the assumption that an approximately invariant space for augmentations is given.  Under that assumption uniformly sampling parameters is sufficient, i.e., there is no need to perform any search for policies or hyperparameters.  
	Indeed, the results presented in the previous section confirm that UA performs on par with existing search-based methods.  This strongly suggests that, for the examined datasets, our assumption holds. 
	We further support the validity of our assumption by running experiments to explore the impact of the ranges of the augmentation operations and NumOps hyperparameter on UA's performance.
	A good augmentation space is one which is aggressive while still keeping the space approximately data invariant.
	To the best of our knowledge, no previous work detailed how they determined the default transforms and their ranges and, with the exception of RA, the same is true for the NumOps parameter.
	Furthermore, while similar experiments would be extremely expensive for the search-based approaches, they are very much practical for UA, given that it is search-free.

	\subsubsection{Investigating the range of magnitudes for augmentation transforms.}
	
	In the experiments above we used the same augmentation operations, listed in Table~\ref{table:augrange}, that were used in other works, e.g., AA, FAA and PBA. 
	The column named ``Default'' in that table shows the values those papers and ours used.  The question we aim to answer now is whether those magnitude ranges yield an approximate invariant augmentation space.  For that we used two different sets of ranges, denoted as ``Narrow'' and ``Wide'' in Table~\ref{table:augrange}; as the name implies they shorten and enlarge, respectively, the range of possible values one can draw from for each augmentation operation.  The hypothesis is
	if the range is too narrow, there will be less diversity in the augmented data, whereas if the ranges are too wide the augmented data would be out-of-distribution.  In both cases this would lead to sub-optimal learning.
	%
	In order to verify the validity of the hypothesis, we trained WideResNet 28x10 on CIFAR-10 dataset with UA using all three magnitude ranges, and compared their performance on the validation set. The results in Table~\ref{table:augrange-results} suggest our hypothesis is correct, and that the ranges of values used in the experiments are appropriate.
	
	\begin{table} [htb]
		\begin{center}
			\caption{List of augmentation image transformations and their ranges. (N/A denotes a binary operation with no magnitude.)}
			\label{table:augrange}
			\begin{tabular}{l | cccccccl}
				\hline
				Transform & Narrow & Default  & Wide \\
				\hline \hline 
				\small ShearX(Y) &  [-0.15,0.15] & [-0.3,0.3] & [-0.9,0.9] \\
				\small TranslateX(Y) & [-0.225,0.225] & [-0.45,0.45] & [-1,+1]\\
				\small Rotate & [-15,15] & [-30,30] & [-90,90] \\
				\small AutoContrast & N/A & N/A & N/A \\
				\small Invert & N/A & N/A & N/A \\
				\small Equalize & N/A & N/A & N/A \\
				\small Solarize & [0,256] & [0,256] & [0,256] \\
				\small Posterize & [6,8] & [4,8] & [2,8] \\
				\small Contrast & [0.5,1.5] & [0.1,1.9] & [0.01,2] \\
				\small Color & [0.5,1.5] & [0.1,1.9] & [0.01,2] \\
				\small Brightness & [0.5,1.5] & [0.1,1.9] & [0.01,2] \\
				\small Sharpness & [0.5,1.5] & [0.1,1.9] & [0.01,2] \\
				\small Cutout & [0,0.1] & [0,0.2] & [0,0.6] \\
				\hline
			\end{tabular}
		\end{center}
	\end{table}
	
	\begin{table} [htb]
		\begin{center}
			\caption{UA's accuracy on CIFAR10 and WideResNet 28x10, using different ranges for the augmentation operations}
			\label{table:augrange-results}
			\begin{tabular}{ccc}
				\hline
				Narrow & Default  & Wide \\
				\hline \hline
				2.77 & 2.67 & 2.99 \\
				\hline
			\end{tabular}
		\end{center}
	\end{table}

	\subsubsection{The effect of NumOps.} 
	
\begin{figure} [htb]
		\centering
		\includegraphics[height=5cm]{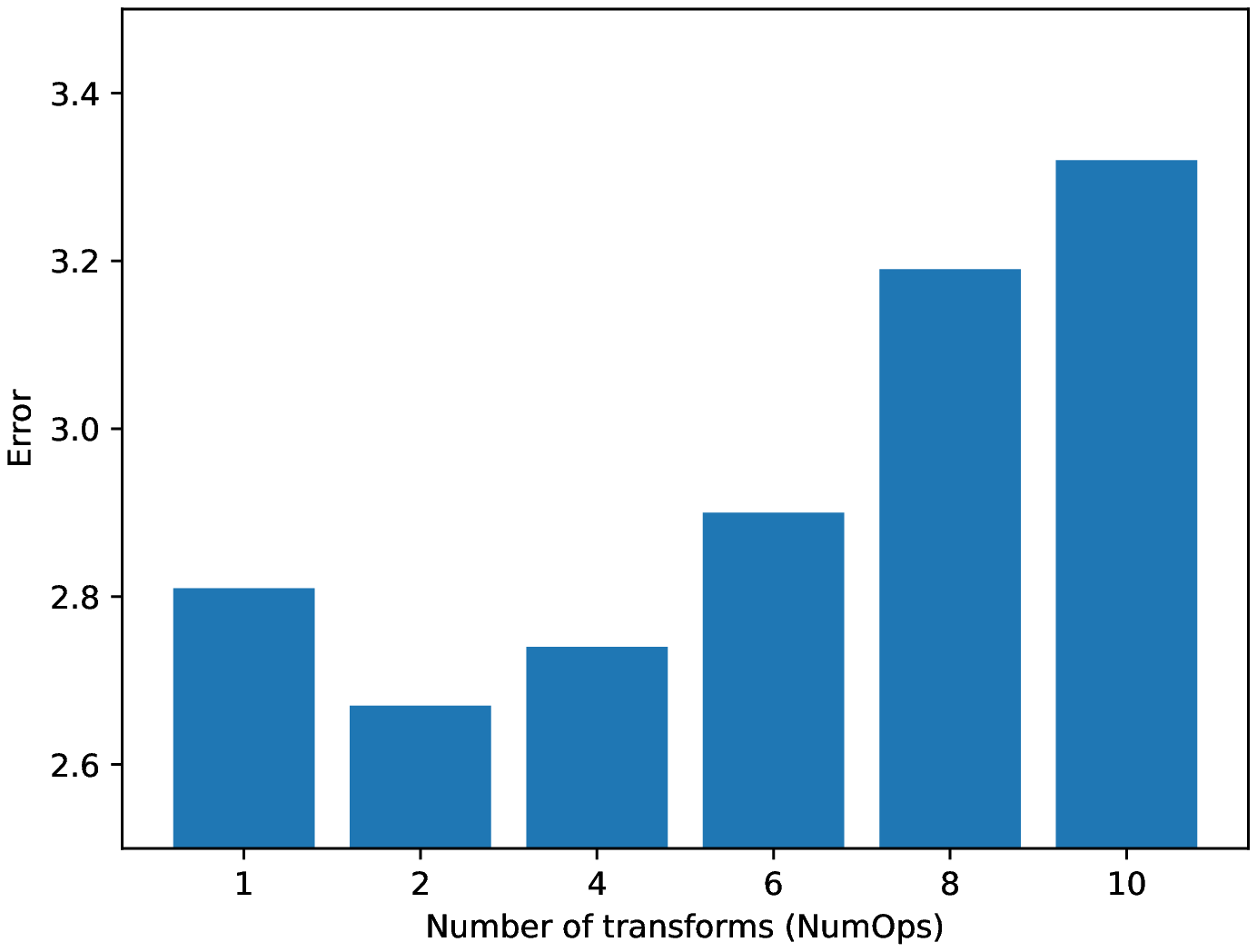}
		\caption{Error rates on CIFAR10 validation set for WideResNet 28x10.}
		\label{fig:numops}
	\end{figure}

	Similar to the experiment above we now seek to confirm that NumOps = 2 yields an approximately invariant augmentation space.
	Again, similarly to the case above, the intuition is that if NumOps = 1, less diversity is added to the learning but if NumOps $\gg$ 1 the augmented image will become out-of-distribution.
	%
	Figure~\ref{fig:numops} shows the error rate on validation set for different values of NumOps of UA on CIFAR10. 
	It confirms the intuition that while a single augmentation does not add much diversity to the data, more than 4 augmentations quickly degenerates the image generating out of distribution samples, and reducing the effectiveness.
	Moreover, it further suggests that choosing NumOps as 2 is indeed a very reasonable choice for the datasets investigated, 
	which is in line with RA's proposal \cite{cubuk2019randaugment}.
	
	The two sets of experiments above clearly support our hypothesis that  the sets of hyperparameters we use form an approximately invariant augmentation space.  
	But it also leads to a more fundamental question: \emph{Can one discover (or design) approximately invariant augmentation space efficiently for a given domain?}  While we defer that for future work, an affirmative answer would be very useful as it would imply that a highly efficient and effective search-free approach, such as UA, could be applied to such domain.

	\section{Conclusions}
	\label{sec:conclusions}
	
	Most of the recently proposed automatic augmentation methods are computationally expensive as they rely on the use of a proxy model and search for the optimum augmentation operations and their parameters (for that given proxy model) for the augmentation operations.
	In this paper we demonstrated that if the augmentation space is approximately invariant, which relatively restricts generating out of distribution samples, the need to search within this space for optimum parameters is eliminated.
	This observation lead to our main contribution, UniformAugment (UA). That is, if the assumption of an approximately invariant augmentation space holds, it suffices to uniformly sample augmentation operations from such a space.  This avoids any search while leading the trained model to deliver performance comparable to the existing search-based methods.  
	
	Even though we cannot guarantee that the default predefined ranges for transform operations used in this and many other papers do yield a approximately invariant augmentation space, the very positive results we obtained seem to suggest as much, for the datasets and models we used.
	Indeed, we argue that an important direction for further research is to develop a methodology to search for an approximately invariant augmentation space for a given task and domain.  This is more important than searching for the optimum parameters within a given augmentation space. If such approximately invariant augmentation space is obtained then no further search is needed, and an approach such as UA, being resilient to different datasets and/or updates to those, is bound to perform well for that task/domain. 
	
	Another interesting direction for further work is to compare UA's effectiveness against approaches using adversarial methods for policy generation \cite{zhang2019adversarial}.  While that approach does not have a proxy-model search \emph{per se}, it requires a more expensive training process. It would be interesting to also deploy UA in a similar setting using batch augmentation\cite{hoffer2019augment}, to investigate its effectiveness with larger augmented batches.

	\section{Acknowledgements}
	The first five authors contributed equally to this work. 
	M. Nascimento's contributed while on sabbatical leave at the Distributed Data Lab, Huawei Technologies, Canada.
	We gratefully acknowledge fruitful discussions with Robin Grosman's as well as her support during the development of this work.

	

	\bibliographystyle{splncs04}
	\bibliography{main}
\end{document}